\documentclass{article}

\usepackage[final]{nips_2018}

\usepackage[utf8]{inputenc} 
\usepackage[T1]{fontenc}    
\usepackage{hyperref}       
\usepackage{url}            
\usepackage{booktabs}       
\usepackage{amsfonts}       
\usepackage{nicefrac}       
\usepackage{microtype}      
\usepackage{amsmath,amssymb,amsthm,mathrsfs,amsfonts,dsfont} 
\usepackage{algorithm}
\usepackage{algpseudocode}
\usepackage{subcaption}

\usepackage[pdftex]{graphicx}

\newcommand{\Lo}{\mathcal{L}}
\newcommand{\1}{\mathds{1}}
\newcommand{\la}{\ell}
\newcommand{\E}{\mathds{E}}
\newcommand{\X}{\mathcal{X}}

\newcommand{\C}{\mathcal{C}}
\newcommand{\Y}{\hat{Y}}

\newcommand{\w}{\mathbf{w}}

\newcommand{\var}{\mathds{V}\mathrm{ar}}

\DeclareMathOperator*{\argmax}{arg\,max}

\newcommand{\dy}{\,\mathrm{d}y}
\newcommand{\dx}{\,\mathrm{d}x}
\newcommand{\x}{\mathbf{x}}

\usepackage{color}
 \usepackage{authblk}

\title{SHADE: Information-Based Regularization for Deep Learning}

\author[]{Michael Blot}
\author[]{Thomas Robert}
\author[]{Nicolas Thome}
\author[]{Matthieu Cord}
\affil[]{Sorbonne Université, CNRS, Laboratoire d’Informatique de Paris 6, LIP6, F-75005 Paris, France}

\begin{document}

\maketitle

\begin{abstract}
    Regularization is a big issue for training deep neural networks. In this paper, we propose a new information-theory-based regularization scheme named SHADE for SHAnnon DEcay. The originality of the approach is to define a prior based on conditional entropy, which explicitly decouples the learning of invariant representations in the regularizer and the learning of correlations between inputs and labels in the data fitting term. Our second contribution is to derive a stochastic version of the regularizer compatible with deep learning, resulting in a tractable training scheme. We empirically validate the efficiency of our approach to improve classification performances compared to common regularization schemes on several standard architectures.
\end{abstract}

\section{Introduction}
    Deep neural networks (DNN) have shown impressive state-of-the-art results in the last years on numerous tasks and especially on image recognition~\cite{alexnet,resnet}. In image classification particularly, deep convolutional neural networks, have demonstrated an outstanding ability to generalize, despite being able to completely learn the training set. 
    Indeed, \cite{rethinking} experimentally demonstrate that generalization performance of DNN remains an open question, that usual machine learning tools, such as VC dimension \cite{vcdim} or Rademacher \cite{rademacher} complexity fail to explain. Further more, many regularization tricks such as weight decay~\cite{weightdecay}, dropout~\cite{dropout} or batch normalization~\cite{batchnorm} seem crucial in the training of DNN, whereas they lake for theoretical foundations enabling to interpret their effect.

    In this article, we study the possibility to design a regularization scheme that can be applied efficiently to deep learning and that has theoretical motivations. Our approach requires to define a regularization criterion, which is our first contribution. We claim that, for any model, the entropy of its intermediate representation, conditionally to the class variable, is a good criterion to apprehend the generalization potential of the model. More formally let's note $X \in \X$ the input variable, $C \in \C$ the class variable, $w$ the model parameters with $Y=h(w,X)$ the (deep) representation of the input leading to the class prediction. Then, the objective is to penalize $H(Y\mid C)$, where $H$ denotes the Shannon entropy measure (see~\cite{element} for definition). As explain in next section, the measure $H(Y\mid C)$ is perfectly suitable to quantify how invariant the representation $Y$ is, accordingly to the underlying task of class prediction. This criterion also stands as a valid instantiate of the "Minimum Description Length Principle"~\cite{mdl}, an interpretation of the Occam Razor. 
    
    Unfortunately, information measures are usually difficult to estimate when the number of data is low, compared to the size of the variable support space. As a second contribution, we propose an implementation of a tractable loss that proves to reduce our criterion when minimized. Indeed, based on an interpretation of the class information encoding within neurons activations, we assume that for every neuron exists a random Bernoulli variable that contains most of the class information. This variable ultimately enables to derive a batch-wise estimator of the entropy criterion, that is scalable and integrates easily in a stochastic gradient descent (SGD) optimization scheme. The resulting loss, called SHADE for SHAnnon DEcay, has the advantage to be layer-wise and more particularly neuron-wise. 
    
    Finally, as a third contribution we provide extensive experiments on different datasets to motivate and to validate our regularization scheme. 
    
\section{Related work and motivation}
    \label{motivation}
    \paragraph*{Regularization in deep learning.}
        For classification tasks, DNN training schemes usually use an objective function which linearly combines a classification loss $\ell_{\mathrm{cls}}(w, X, C)$ --\,generally cross-entropy\,--  and a regularization term $\Omega(w, X, C)$, with $\beta \in \mathbb{R}^+$, that aims at influencing the optimization toward a local minima with good generalization properties:
        \begin{equation}
        \label{eq:loss}
            \Lo(w) = \E_{(X,C)} \big[ \ell_{\mathrm{cls}}(w, X, C) + \beta\cdot\Omega(w, X, C) \big]
        \end{equation}
        For example, the weight decay (WD) loss~\cite{weightdecay} is supposed to penalize the complexity of the model. While there is strong motivations to WD use on linear models, in term of margins or in term of Lipschitz coefficient for instance, the extension of those theoretical results to DDN is not straightforward and the effects of WD on DNN's generalization performances is still not clear as demonstrated in~\cite{rethinking}.
        
        Our SHADE regularization scheme belongs to this family as we construct a loss $\Omega_{SHADE}(h(X, w), C)$, that aims at influencing the optimization toward representations with low class conditioned entropy. We show in the experiments that SHADE loss has a positive effect on our theoretical criterion $H(Y\mid C)$, resulting in enhanced generalization performances. 
        
        Others popular regularization methods like~\cite{dropout, conf/icml/WanZZLF13} deactivate part of the neurons during the training. Those methods, which tend to lower the dependency of the class prediction to a reduced set of features, is backed by some theoretical interpretations like~\cite{2016arXiv161101353A, bayes}. Other methods that add stochasticity to the training such as batch normalization or stochastic  poolings~\cite{batchnorm, zeiler2013stochastic, graham2014fractional}, beside being comparable to data-augmentation, result in the addition of noise to the gradients during optimization. This property would enable the model to converge toward a flatter minima, that are known to generalize well as shown in~\cite{largebatch}. More generally, stochastic methods tend to make the learned parameters less dependant on the training data, which guarantee tighter generalization bounds~\cite{nipsInformationGeneralization}. 
        In this article we focus on another way to make DNN's models less dependant on the training data. By focusing on representation that are invariant to many transformations on the input variable, you make the training less dependant on the data. Having invariant representation is the motivation for our criterion $H(Y\mid C)$.
        
    \paragraph*{Quantifying invariance.} 
        Designing DNN models that are robust to variations on the training data and that preserve class information is the main motivation of this work. In the same direction, Scattering decompositions~\cite{scattering} are appealing transforms. They have been incorporated into adapted network architectures like~\cite{bruna}. However, for tasks like image recognition, it is very difficult to design an explicit modelling of all transformations a model should be invariant to. 
        
        Inversely, a criterion such as $H(Y\mid C)$ is agnostic to the transformations the representation should be invariant to, and is suitable to quantify how invariant it is in a context of class prediction. Indeed, a model that is invariant to many transformations will produce the same representation for different inputs, making it impossible to guess which input produced a given representation. This characteristic is perfectly captured by the entropy $H(X\mid Y)$ which represents the uncertainty in the variable $X$ knowing its representation $Y=h(w,X)$. For instance, many works relate the reconstruction error to the entropy $H(X\mid Y)$ like in Fano's inequality \cite{element} in the discrete case. A general discussion with illustrations on how to bound the reconstruction error with $H(X\mid Y)$ can be found in Appendix \ref{invariantentropy}. Thus, the bigger the measure $H(X\mid Y)$, the more invariant the representation. Let's now analyze the entropy of the representation $Y$ when $X$ is discrete and for a deterministic mapping $Y=h(w,X)$. We have:
        \begin{equation}
        H(Y) = I(X,Y) + \underbrace{H(Y \mid X)}_{=0\text{ (determ.)}} = I(X, Y) = H(X) - H(X \mid Y).
        \end{equation}
        $H(X)$ being fixed, $H(Y)$ is inversely related to $H(X\mid Y)$ making $H(Y)$ also a good measure of invariance. 
        
        However, considering the target classification task, we do not want two inputs of different classes to have the same representation but rather want to focus on intra-class invariance. Therefore, all this reasoning should be done conditionally to the class $C$, explaining final choice of $H(Y\mid C)$\footnote{$H(Y\mid C) = H(X\mid C) - H(X\mid Y,C)$} as a measure of intra-class invariance.

    \paragraph*{Information-theory-based regularization.}
        Many works like~\cite{DBLP:journals/corr/PereyraTCKH17, soatto} use information measures as regularization criterion. Still with the objective of making the trained model less dependant on the data, \cite{soatto} has built a specific weight regularization but had to model the weight distribution which is not easy. The information criterion that is closer to ours is the one defined in the Information Bottleneck framework (IB) proposed in~\cite{IB} that suggests to use mutual information $I(X,Y)$ (\cite{element})\footnote{for all variable $X$, $Y$, $I(X,Y) = H(X) - H(X|Y)  = H(Y) - H(Y|X)$} as a regularization criterion. \cite{IBvariational} extends it in a variational context, VIB, by constructing a variational upper-bound of the criterion. Along with IB also come some theoretical investigations, with the definition of generalization bounds in~\cite{IBbound}. Using mutual information as a regularizer is also closely connected to invariance since $I(X,Y)$ attempts at compressing as much information as possible from input data. 
        In the case $X$ is discrete and the representation mapping is deterministic ($H(Y\mid X)=0$), our criterion is related to IB's trough the following development $I(X, Y) = H(Y) = I(C, Y) + H(Y\mid C)$. In a context of optimization with SGD, minimizing $H(Y\mid C)$ appears to be more efficient to preserve the term $I(C, Y)$, which represents the mutual information of the representation variable with the class variable and which must stay high to predict accurately $C$ from $Y$. 
        
        Compressing the representation, without damaging the class information seems in fact to be a holy grail in machine learning. Our work, resulting in SHADE, goes in this direction.

\section{SHADE: A new Regularization Method Based on \texorpdfstring{$H(Y\mid C)$}{H(Y|X)}}
\label{sec:SHADE} 
    In this section, we will further describe SHADE, a new regularization loss based on the conditional entropy $H(Y\mid C)$ designed to drive the optimization toward more invariant representation. We first show how to derive a layer-wise, neuron-wise criterion before developing a proper tractable estimate of the unit-wise criterion. In order to properly develop entropy inequalities it is necessary to suppose that $X$ is a discrete variable of the space $ \{0,...,255\}^{H \times W \times 3}$ with $H$ and $W$ respectively the height and width of the images. However it is common to consider $X$ as a discrete quantization of a continuous natural variable, enabling to exploit some properties verified by continuous variables such as gradient computing\footnote{A discussion on this topic can be found in~\cite{discussionIB}}.

    \subsection{A unit-wise criterion}
    
        \begin{figure}
            \centering
             \includegraphics[width=0.8\textwidth]{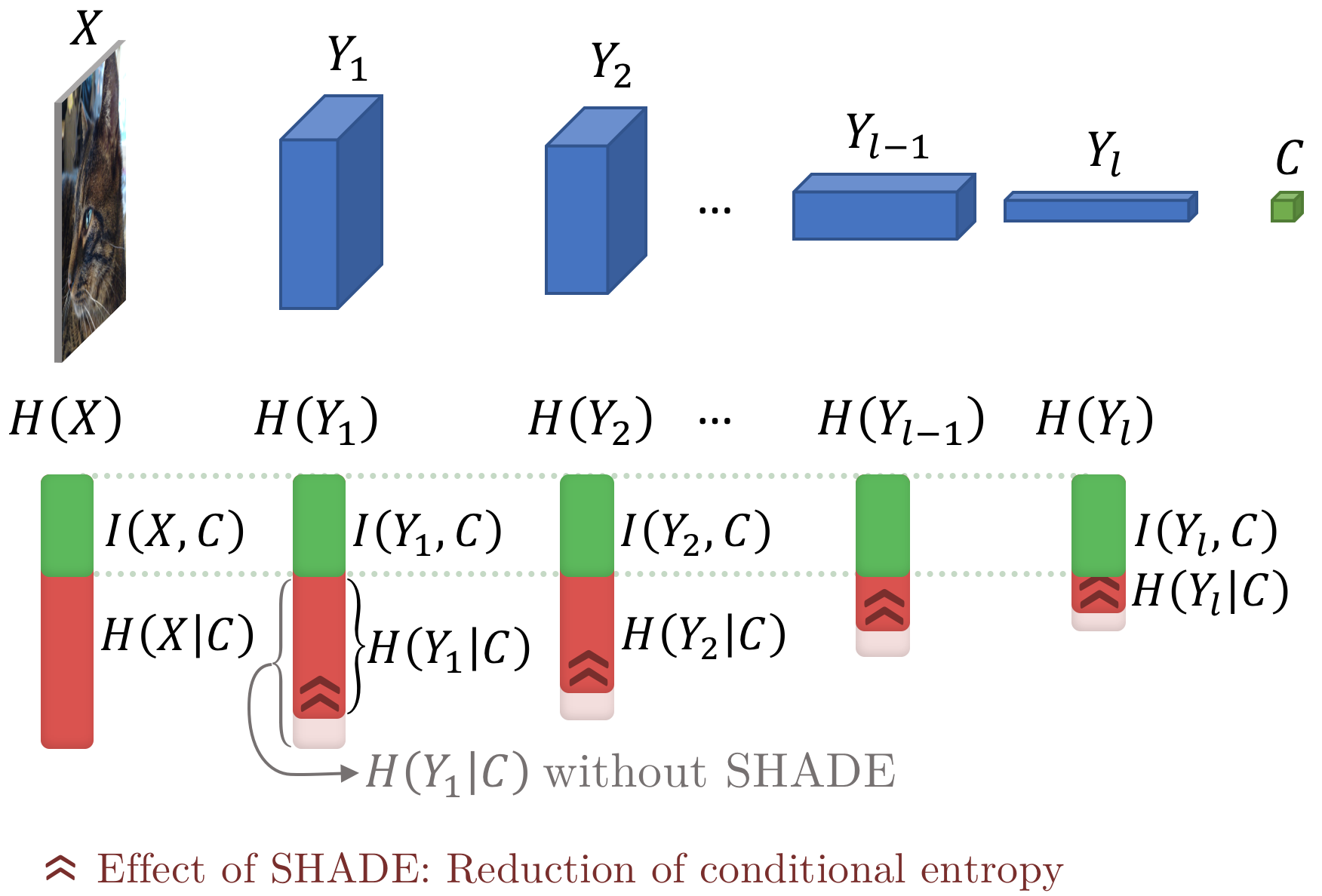}
            \caption{DNN architecture with corresponding layers' entropies, showing the layer-wise action of SHADE. Given that $H(Y_i) = I(Y_i,C)+H(Y_i\mid C)$, SHADE minimizes $H(Y_i\mid C)$ without affecting $I(Y_i,C)$.}
            \label{fig:layer-wize-regularization}
        \end{figure}
    
        \paragraph{Layer-wise criterion.} A DNN is composed of a number $L$ of layers that sequentially transform the input. Each one can be seen as an intermediate representation variable, noted $Y_\la$ for layer $\la \in \{1,..,L \}$, that is determined by the output of the previous layer $Y_{\la-1}$ and a set of parameters $\w_\la$. Each layer filters out a certain part of the information from the initial input $X$. Thus, the following inequalities can be derived from the data processing inequality in~\cite{element}:
        \begin{equation}
            H(Y_L \mid C) \le H(Y_{L-1}\mid C) \le ... \le  H(Y_{1}\mid C) \le H(X\mid C).
        \end{equation}

        The conditional entropy of every layer is upper bounded by the conditional entropy of the previous layer. Similarly to the recommendation of~\cite{IBdeep}, we apply this regularization to all the layers, using a layer-wise criterion $H(Y_\la \mid C)$, and producing a global criterion\footnote{Confirming the intuition, in our experiments, regularizing all layer has proved to be more efficient than regularizing the final layer representation only}:
        \begin{equation}
            \Omega_{\mathrm{layers}} = \sum_{\la = 1}^L \beta_\la H(Y_\la \mid C).
        \end{equation}
        Where $\beta_\la$ is a weighting term differentiating the importance of the regularization between layers. Those coefficient will be omitted in the following as in our experiments all $\beta_\la$ are identical. Adjusting the values of the variables $\beta_\la$ remains open for further research. It is illustrated in Fig.~\ref{fig:layer-wize-regularization} where we see that $I(Y_l,C)$ (in green) remains constant while $H(Y_l\mid C)$ (in red) decreases.
        
        \paragraph{Unit-wise criterion.} Examining one layer $\la$, its representation variable is a random vector of coordinates $Y_{\la,i}$ and of dimension $D_\la$: $Y_\la = (Y_{\la,1},..., Y_{\la, D_\la})$. The upper bound 
        $H(Y_\la \mid C) \le \sum_{i=1}^{D_\la} H(Y_{\la,i} \mid C)$ enables to define a unit-wise criterion that SHADE seeks to minimize. For each unit $i$ of every layer $\la$ we design a loss $\Omega_{\mathrm{unit}}(Y_{\la,i} \mid C)=H(Y_{\la,i} \mid C)$ that will be part of the global regularization loss:
            \begin{align}
            \label{SHADE}
                \Omega_{\mathrm{layers}} \le \sum_{l=1}^L \sum_{i = 1}^{D_\la} \underbrace{H(Y_{\la,i} \mid C)}_{\Omega_{\mathrm{unit}}(Y_{\la,i}\mid C)}.
            \end{align}
        
        For the rest of the paper, we use the notation $Y$ instead of $Y_{\la,i}$ for simplicity, since the layers and coordinates are all considered independently to define $\Omega_{\mathrm{unit}}(Y_{\la,i}\mid C)$. 
        
    \subsection{Estimating the Entropy}
        In this section, we describe how to define a loss based on the measure $H(Y\mid C)$ with $Y$ being one coordinate variable of one layer output. Defining this loss is not obvious as the gradient of $H(Y\mid C)$ with respect to the layer's parameters may be computationally intractable. $Y$ has an unknown distribution and without modeling it properly it is not possible to compute $H(Y\mid C)$. Since $H(Y\mid C) = \sum_{c \in \C}p(c)H(Y\mid c)$, a direct approach would consist in computing $|\C|$ different entropies $H(Y\mid c),\ 1\le c\le |\C|$. This means that, given a batch, the number of samples used to estimate one of these entropies is divided by $|\C|$ on average which becomes particularly problematic when dealing with a large number of classes such as the 1,000 classes of ImageNet. Furthermore, entropy estimators are extremely inaccurate considering the number of samples in a batch. For example, LME estimators of entropy in~\cite{entropyestimation} converge in $\mathcal{O}(\nicefrac{(\log K)^2}{K})$ for $K$ samples. Finally, most estimators require to discretise the space in order to approximate the distribution \textit{via} a histogram. This raises issues on the bins definition considering that the variable distribution is unknown and varies during the training in addition to the fact that having a histogram for each neuron is computationally and memory consuming. Moreover, entropy estimators using kernel density estimation usually have a too high complexity ($\approx \mathcal{O}(K^2)$) to be applied efficient on deep learning models.  
        
        To tackle these drawbacks we propose the two following solutions: the introduction of a latent variable that enables to use more examples to estimate the conditional entropies; and a bound on the entropy of the variable by an increasing function of its variance to avoid the issue of entropy estimation with a histogram, making the computation tractable and scalable.
    
        \paragraph{Latent code.}
            First, considering a single neuron $Y$ (before ReLU), the ReLU activation induces a detector behavior toward a certain pattern. If the pattern is absent from the input, the signal is zero; if it is present, the activation quantifies the resemblance with the pattern.
            We therefore propose to associate a binomial variable $Z$ with each unit variable $Y$ (before ReLU). This variable $Z$ indicates if a particular pattern is present on the input ($Z=1$ when $Y \gg 0$) or not ($Z=0$ when $Y \ll 0$).
            It acts like a latent code from which the input $X$ is generated like in variational models (\textit{e.g. }\cite{aevb}) or in generative models (\textit{e.g.} \cite{infogan}).
            
            Furthermore, it is very likely that most intermediate features of a DNN can take similar values for inputs of different classes -- this is especially true for low-level features. The semantic information provided by a feature is thus more about a particular pattern than about the class itself. Only the association of features allows determining the class. So $Z$ represents a semantically meaningful factor about the class $C$. The feature value $Y$ is then a quantification of the possibility for this semantic attribute $Z$ to be present in the input or not.
            
            We thus assume the Markov chain $C \rightarrow Z \rightarrow X \rightarrow Y$. Indeed, during the training, the distribution of $Y$ varies in order to get as close as possible to a sufficient statistic of $X$ for $C$ (see definition in \cite{element}). Therefore, we expect $Z$ to be such that $Y$ draws near a sufficient statistic of $Z$ for $C$ as well. By assuming the sufficient statistic relation $I(Y, C) = I(Y, Z)$ we get the equivalent equality $H(Y\mid C)= H(Y\mid Z)$, and finally obtain:
            \begin{equation}
                H(Y\mid C)= H(Y\mid Z) = \sum_{z\in\{0,1\}} p(z) H(Y\mid Z=z).
            \end{equation}
            
            This modeling of $Z$ as a Bernoulli variable (one for each unit) has the advantage of enabling good estimators of conditional entropy since we only divide the batch into two sets for the estimation ($z=0$ and $z=1$) regardless of the number of classes. The fact that most of $Y$ information about $C$ is contained in such a variable $Z$ is validated in the experiments Sec. \ref{sec:furtherExpe}.

        \paragraph{Variance bound.}
            The previous trick allows computing fewer entropy estimates to obtain the global conditional entropy, therefore increasing the sample size used for each entropy estimation. Unfortunately, it does not solve the bin definition issue. To address this, we propose to use the following bound on $H(Y\mid Z)$, that does not require the definition of bins:
            
            \begin{equation}
                \label{variancebound}
                H(Y \mid Z) \le \frac{1}{2}\ln\big(2 \pi e \var(Y\mid Z)\big).
            \end{equation}
            
            This bound holds for any continuous distributions $Y$ and there is equality if the distribution is Gaussian. For many other distributions such as the exponential one, the entropy is also equal to an increasing function of the variance. In addition, one main advantage is that variance estimators are much more robust than entropy estimators, converging in $\mathcal{O}(\nicefrac{1}{K})$ for $K$ samples instead of $\mathcal{O}(\nicefrac{\log(K)^2}{K})$. The use of this bound is well justified in our case because the variable $Y$ is the quantization of a continuous variable. Moreover, even if $Y$ is discrete, this inequality still holds with respect to a term depending on the quantization steps.  
            
            The $\ln$ function being one-to-one and increasing, we only keep the simpler term $\var(Y\mid Z)$ to design our final loss:
            \begin{equation}
                \Omega_\mathrm{SHADE} = \sum_{\la=1}^{L}\sum_{i=1}^{D_\la}\sum_{z \in \{0,1\}} p(Z_{\la,i} = z \mid Y) \var(Y\mid Z_{\la,i}=z).
            \end{equation}
            
            In the next section, we detail the definition of the differential loss, computed on a mini-batch, using $\var(Y\mid Z)$ as a criterion. 
            
        \subsection{Instantiating SHADE}   
        \label{instance}
            \begin{algorithm}[tb]
                    \caption{Moving average updates:
                    for $z \in \{0,1 \}$, $p^z$ estimates $p(Z = z)$ and $\mu^z$ estimates $\E(Y\mid Z = z)$}
                    \label{alg:maupdate}
                    \begin{algorithmic}[1]
                    \State \textbf{Initialize:} $\mu^0 = -1$, $\mu^1 = 1$, $p^0 = p^1 = 0.5$, $\lambda=0.8$             
                    \renewcommand{\algorithmicforall}{\textbf{for each}}
                    \ForAll{mini-batch $\{y^{(k)}, k \in 1 .. K\}$}
                        \For{$z \in \{ 0,1 \}$}
                            \State $p^{z} \leftarrow  \lambda p^{z} +   (1- \lambda)\frac{1}{K}\sum_{k=1}^K p(z\mid y^{(k)})$
                            \State $\mu^{z} \leftarrow \lambda  \mu^{z} +  (1-\lambda)\frac{1}{K}\sum_{k=1}^K\displaystyle \frac{p(z\mid y^{(k)})}{p^z}y^{(k)}$
                        \EndFor
                    \EndFor
                \end{algorithmic}
            \end{algorithm}

            For one unit of one layer, the previous criterion writes:
            \begin{align}
            \label{varianceDev}
                \var(Y \mid Z) &= \int_{\Y} p(y) \sum_{z \in \{0,1\}}p(z\mid y)\big(y-\E(Y\mid z)\big)^2 \dy \\
                &\approx \frac 1 K \sum_{k=1}^K \left[\sum_{z \in \{0,1\}}p(z\mid y^{(k)})\big(y^{(k)}-\E(Y\mid z)\big)^2\right].
                \label{eq:MC}
            \end{align}
            
            The quantity $\var(Y \mid Z)$ can be estimated with Monte-Carlo sampling on a mini-batch of $K$ input-target pairs $\big\{(x^{(k)}, c^{(k)})\big\}_{1 \le k  \le K}$ of intermediate representations $\big\{y^{(k)}\big\}_{1 \le k \le K}$ as in Eq.~(\ref{eq:MC}).

            $p(Z \mid y)$ is interpreted as the probability of presence of attribute $Z$ on the input, and should clearly be modeled such that $p(Z = 1 \mid y)$ increases with $y$. The more similarities between the input and the pattern represented by $y$, the higher the probability of presence for $Z$. We suggest using\footnote{Other functions have been experimented with similar results}:
            \begin{equation*}
                p(Z=1\mid y) = 1-e^{-ReLU(y)} \qquad p(z=0 \mid y) = e^{-ReLU(y)}.
            \end{equation*}
            
            For the expected values $\mu^z = \E(Y\mid z)$  we use a classic moving average that is updated after each batch as described in Algorithm \ref{alg:maupdate}. Note that the expected values are not changed by the optimization since translating a variable has no influence on its entropy.
            
            The concrete behavior of SHADE can be interpreted by analyzing its gradient as described in Appendix \ref{sec:gradients}.
        
            For this proposed instantiation, our SHADE regularization penalty takes the form:
            \begin{equation}
                \Omega_{\mathrm{SHADE}} = \sum_{\la=1}^{L}\sum_{i=1}^{D_\la}\sum_{k=1}^K \sum_{z \in \{0,1\}} p\left(Z_{\la,i} = z\,\middle|\, y_{\la,i}^{(k)}\right) 
                \left({y_{\la,i}^{(k)}} - \mu_{\la,i}^ z\right)^2.
            \end{equation}
    
    We have presented a regularizer that is applied neuron-wise and that can be integrated into the usual optimization process of a DNN. The additional computation and memory usage induced by SHADE is almost negligible (computation and storage of two moving averages per neuron). For comparison, SHADE adds half as many parameters as batch normalization does. 

\section{Experiments}
\label{sec:expes}
    \subsection{Image Classification with Various Architectures on CIFAR-10}
    \label{sec:cifarexpe}
    
    \begin{table}[h]
            \caption{Classification accuracy (\%) on CIFAR-10 test set.}
            \label{accuracies}
                \centering
                \begin{tabular}{lccccc}
                \toprule
                & MLP & AlexNet & Inception & ResNet\\
                \midrule
                No regul.        & 64.68 & 83.25 & 91.21 & 92.95 \\
                Weight decay     & 66.52 & 83.54 & 92.87 & 93.84\\
                Dropout          & 66.70 & 85.95 & 91.34 & 93.31\\
                \cmidrule{1-5}
                SHADE            & \textbf{70.45} & \textbf{85.96} & \textbf{93.56} &\textbf{93.87}\\
                \bottomrule
                \end{tabular}
        \end{table}
         
        We perform image classification on the CIFAR-10 dataset, which contains 50k training images and 10k test images of 32$\times$32 RGB pixels, fairly distributed within 10 classes (see \cite{cifar} for details). Following the architectures used in~\cite{rethinking}, we use a small Inception model, a three-layer MLP, and an AlexNet-like model with 3 convolutional layers (64 filters of size $3\times 3$) + max pooling and 2 fully connected layers with 1000 neurons for the intermediate variable. We also use a ResNet architecture from~\cite{wideresnet} (k=10, N=4). Those architectures represent a large family of DNN and some have been well studied in~\cite{rethinking} within the generalization scope. For training, we use randomly cropped images of size 28$\times$28 with random horizontal flips. For testing, we simply center-crop 28$\times$28 images. We use momentum SGD for optimization (same protocol as \cite{rethinking}).
           
        We compare SHADE with two regularization methods, namely {\em weight decay} \cite{weightdecay} and {\em dropout} \cite{dropout}. 
        For all architectures, the regularization parameters have been cross-validated to find the best ones for each method and the obtained accuracies on the test set are reported in Table~\ref{accuracies}. Find more details on the experiments protocol in \ref{sec:cifar}.
        
        We obtain the same trends as~\cite{rethinking}, which shows a small improvement of 0.29\% with weight decay on AlexNet. The improvement with weight decay is slightly more important with ResNet and Inception (0.79\% and 1.66\%). 
        In our experiments dropout improves significantly generalization performances only for AlexNet and MLP. It is known that the use of batch normalization and only one fully connected layer lowers the benefit of dropout, which is in fact not used in~\cite{resnet}.
        
        We first notice that for all kind of architectures that the use of SHADE significantly improves the generalization performances, 5.77\% for MLP, 2.71\% for AlexNet, 2.35\% for Inception and 0.92\% for ResNet. It demonstrates the ability of SHADE to regularize the training of deep architectures. 
        Finally, SHADE shows better performances than dropout and weight decay on all architectures.
            
    \subsection{Large Scale Classification on ImageNet}
    \label{imagenetExperiment}
        In order to experiment SHADE regularization on a very large scale dataset, we train on ImageNet~\cite{ImageNet} a WELDON network from~\cite{weldone} adapted from ResNet-101. This architecture changes the forward and pooling strategy by using the network in a fully-convolutional way and adding a max+min pooling, thus improving the performance of the baseline network.
        We used the \textbf{pre-trained weights of ResNet-101} (from the torchvision package of PyTorch) giving performances on the test set of \textbf{77.56\%} for top-1 accuracy and 93.89\% for top-5 accuracy. Provided with the pre-trained weights, the \textbf{WELDON architecture} obtains \textbf{78.51\%} for top-1 accuracy and 94.65\% for top-5 accuracy. After fine tuning the network using \textbf{SHADE} for regularization we finally obtained \textbf{80.14\%} for top-1 accuracy and 95.35\% for top-5 accuracy for a concrete improvement. This demonstrates the ability to apply SHADE on very large scale image classification successfully.

    \subsection{Training with a Limited Number of Samples}
        
        \begin{figure}[tb]
            \centering
            \begin{subfigure}[t]{0.49\textwidth}
                \includegraphics[width=\textwidth]{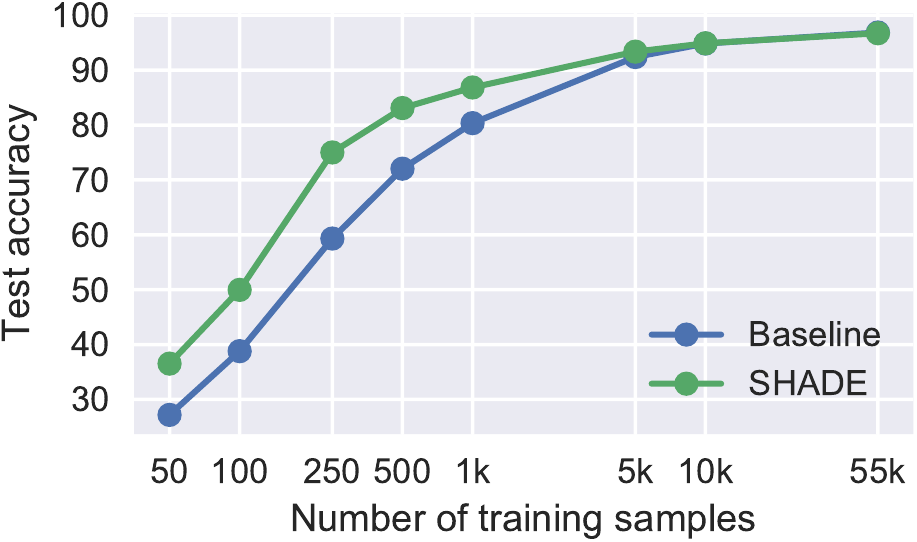}
                \caption{Results for MNIST-M}
                \label{fig:limited_mnistm}
            \end{subfigure}%
            ~
            \begin{subfigure}[t]{0.49\textwidth}
                 \includegraphics[width=\textwidth]{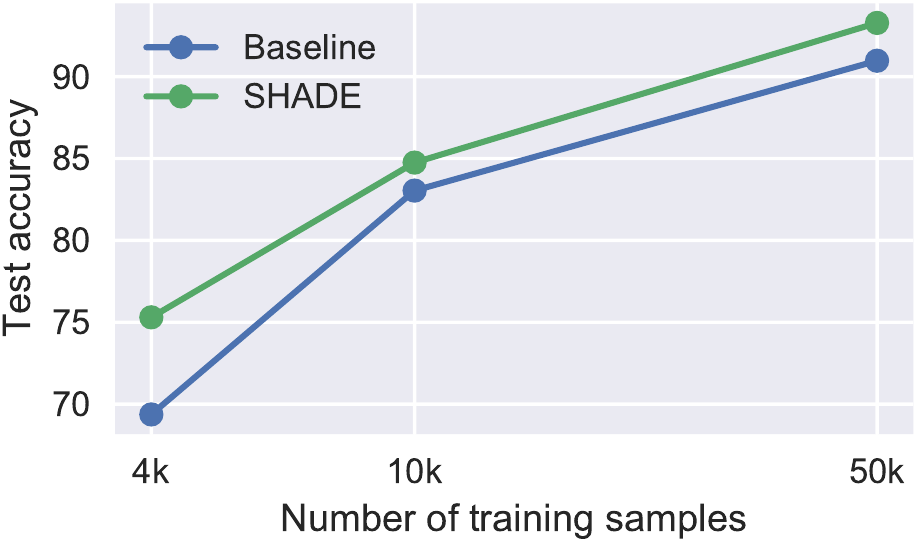}
                \caption{Results for CIFAR-10}
                \label{fig:limited_cifar10}
            \end{subfigure}
            
            \vspace{5mm}
            \begin{subfigure}[t]{0.85\textwidth}
                \centering
                \includegraphics[width=\textwidth]{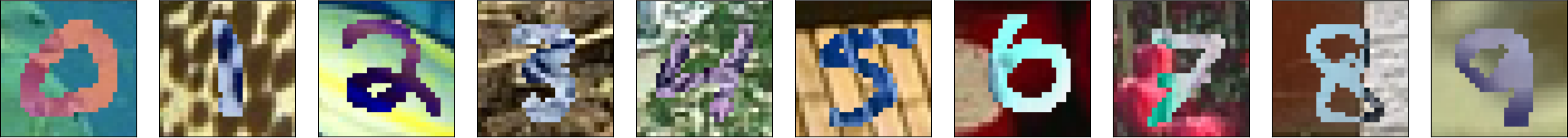}
                \caption{Examples of MNIST-M images misclassified by the baseline and correctly classified using SHADE, both trained with 250 samples.}
                \label{fig:limited_mnistm_viz}
            \end{subfigure}
            
            \caption{Results when training with a limited number of samples in the training set for MNIST-M and CIFAR-10 with and without SHADE.}
            \label{fig:limited_dataset}
        \end{figure}
        
        When datasets are small, DNN tend to overfit quickly and regularization becomes essential. Because it tends to filter out information and make the network more invariant, SHADE seems to be well fitted for this task. To investigate this, we propose to train DNN with and without SHADE on CIFAR-10 and MNIST-M with different numbers of samples in the training set.
        
        First, we tested this approach on the digits dataset MNIST-M~\cite{ganin2015unsupervised}. This dataset consists of the MNIST digits where the background and digit have been replaced by colored and textured information (see Fig. \ref{fig:limited_mnistm_viz} for examples). The interest of this dataset is that it contains lots of unnecessary information that should be filtered out, and is therefore well adapted to measure the effect of SHADE. A simple convolutional network has been trained with different numbers of samples of MNIST-M and the optimal regularization weight for SHADE have been determined on the validation set (see training details in Appendix \ref{sec:mnistm}). The results can be seen on Figure~\ref{fig:limited_mnistm}. We can see that especially for small numbers of training samples ($<$ 1000), SHADE provides an important gain of 10 to 15\% points over the baseline. This shows that SHADE helped the model in finding invariant and discriminative patterns using less data samples.
        
        Additionally, Figure~\ref{fig:limited_mnistm_viz} shows samples that are misclassified by the baseline model but correctly classified when using SHADE. These images contain a large amount of intra-class variance (color, texture, etc.) that is not useful for the classification tasks, explaining why adding SHADE, that encourages the model to discard information, allows important performance gains on this dataset and especially when only few training samples are given.
        
        Finally, to confirm this behavior, we also applied the same procedure in a more conventional setting by training an Inception model on CIFAR-10. Figure~\ref{fig:limited_cifar10} shows the results in that case. We can see that once again SHADE helps the model gain in performance and that this behavior is more noticeable when the number of samples is limited, allowing a gain of 6\% when using 4000 samples.

    \subsection{Further experiments: exploration on the latent variable}
    \label{sec:furtherExpe}
    
            
            
    
        \begin{table}[tb]
            \caption{Classification accuracy (\%) on CIFAR-10 test set with binarized activation.}
            \label{tab:binaryactivationaccuracy}
                \centering
                \begin{tabular}{c|c}
                \toprule
                MLP & \begin{tabular}{cccc}
                    baseline & layer 3 & layer 2 & layer 1\\
                    64.68 & 64.92 & 62.45 & 61.13                    
                        \end{tabular}\\
                \midrule
                AlexNet & \begin{tabular}{cccc}
                    baseline & last 4 & layer 2 & layer 1\\
                    83.25 & 82.71 & 82.38 & 82.01
                        \end{tabular}\\
                \midrule
                Inception & \begin{tabular}{cccc}
                    baseline & layer 21 & layer 10 & layer 1\\
                    91.34 & 91.41 & 90.88 & 90.21 
                        \end{tabular}  \\      
                \midrule
                ResNet  & \begin{tabular}{cccc}
                    baseline & layer 56 & layer 25 & layer 1\\
                    93.24 & 92.67 & 92.09 & 91.99 
                        \end{tabular} \\
                \bottomrule
                \end{tabular}
        \end{table}
    
        SHADE is based on the intuition that the class information encoded within a neuron is mostly contained in a binary latent variable noted $Z$. To justify this assumption we propose an experiments that studies trained networks neuron variables. 
        In this experiment we work on the possibility to transform the ReLU activation function of a layer into a binary activation function that can only take two values. By exhibiting such a binary activation which does not affect the accuracy, we show that we can summarize the class information of a neuron into a binary variable and still get the same prediction accuracy as with the continuous ReLU activation. The experiment have been done on CIFAR-10 dataset with the same networks used in Sec.~\ref{sec:cifarexpe}.

        \paragraph{Binary activation.} 
        We have replaced the ReLU activation function on a chosen layer of a trained network with a binary activation function. The binary function is $\overline{Y^+}\1(Y \ge \overline{Y^+})$ where $\overline{Y^+}$ stands for the average value of the \textit{positive} variable values before any activation function. After replacing the activation function we fine tune the layers on the top of the chosen layer, in order to adapt the top of the network to the new values and we report the obtained accuracies on the Table~ \ref{tab:binaryactivationaccuracy} for the different networks and different layers. We note that the differences in accuracy are very small losses
        . This means that for a given layer, the information that is used for the class prediction can be sum up in a binary variable confirming the existence of a binary latent variable containing most of the class information that is exploited by the rest of the network. The fact that this apply for all layers of the network is consistent with the application of SHADE loss for all layers. Note that this binary activation could be further researched to improve the modeling integrated in SHADE.

\section{Conclusion}
    In this paper, we introduced a new regularization method for DNN training, SHADE, which focuses on minimizing the entropy of the representation conditionally to the labels. This regularization aims at increasing the intra-class invariance of the model while keeping class information. SHADE is tractable, adding only a small computational overhead when included into an efficient SGD training. We show that our SHADE regularization method significantly outperforms standard approaches such as weight decay or dropout with various DNN architectures on CIFAR-10. We also validate the scalability of SHADE by applying it on ImageNet. The invariance potential brought out by SHADE is further illustrated by its ability to ignore irrelevant visual information (texture, color) on MNIST-M. Finally, we also highlight the increasing benefit of our regularizer when the number of training examples becomes small. Furthermore there is no doubt that the information-theory-based interpretation of SHADE, from which it has been established, allows for further improvements of SHADE for future work.

\bibliographystyle{plain}
\bibliography{biblio}

\appendix

            
            
            
            

\section{Experiments details on CIFAR-10}
    \label{sec:cifar}
    \paragraph{Optimization.} For all experiments the learning rate is initialize and a multiplicative decay is apply on it after every batches. The momentum is constant and setted to 0.9. 
    We detail here the initial learning rate and the decay for every networks used in the format (initial learning rate value, decay): mlp (,), alexnet (,), inception (,), resnet(,)
    \paragraph{Hyperparameter tuning} For weight decay and SHADE, the optimal regularization weight of each model has been chosen to maximize the accuracy on the validation sets. We tried the values $\{[1,5].10^{-i}, i=1..7\}$. For the dropout we have apply it on the two last layer of every networks. The optimal activation probabilities for each model has been chosen among $\{(n/10,m/10), n=1..7,\ m=3..10\}$ to maximize the accuracy on the validation sets.

\section{Experiments details on MNIST-M}
    \label{sec:mnistm}
    \paragraph{Dataset splits and creation.} To create MNIST-M, we kept the provided splits of MNIST, so we have 55,000 training samples, 5,000 validation samples, and 10,000 test samples. Each digit of MNIST is processed to add color and texture by taking a crop in images from BST dataset. This procedure is explained in \cite{ganin2015unsupervised}.
    
    \paragraph{Subsets of limited size.} To create the training sets of limited size $N$, we keep $N/10$ (since there are 10 classes) randomly picked samples from each class. When increasing $N$ we keep the previously picked samples so the training samples for $N=100$ are a subset of the ones for $N=250$. The samples chosen for a given value of $N$ are the same across all models trained using this number of samples.
    
    \paragraph{Image preprocessing.} The only preprocessing applied to the input images is that their values are rescaled from $[0,1]$ to $[-1, 1]$.
    
    \paragraph{Optimization.} For the training, we use mini-batch of size 50 and use Adam optimizer with the recommended parameters, \textit{i.e.} $\lambda_r=0.001, \beta_1=0.9, \beta_2=0.999, \epsilon=10^{-8}$.
    
    \paragraph{Hyperparameter tuning.} For weight decay and SHADE, the optimal regularization weight of each model (for each value of $N$) has been chosen to maximize the accuracy on the validation sets. We tried the values $\{10^{-i}, i=1..7\}$.
    
    \paragraph{Model architecture.} The model have the following architecture:
    
    \begin{itemize}
        \item 2D convolution ($64 \times 5\times 5$ kernel, padding 2, stride 1) + ReLU
        \item MaxPooling $2\times 2$
        \item 2D convolution ($64 \times 3 \times 3$ kernel, padding 1, stride 1) + ReLU
        \item MaxPooling $2\times 2$
        \item 2D convolution ($64 \times 3\times 3$ kernel, padding 1, stride 1) + ReLU
        \item MaxPooling $2\times 2$
        \item Fully connected (1024 inputs, 10 outputs) + SoftMax
    \end{itemize}
    
\section{Experiments details on Imagenet}
    The fine tuning in the experiment section \ref{imagenetExperiment} has been done with momentum-SGD with a learning rate of $10^{-5}$ and a momentum of $0.9$ and a batch size of 16 images. It took 8 epochs to converge.

\section{Entropy bounding the reconstruction error}
\label{invariantentropy}

    In section \ref{motivation} we highlight a link between the entropy $H(X\mid Y)$ and the difficulty to recover the input $X$ from its representation $Y$. Here we exhibit a concrete relation between the reconstruction error, that quantifies the error made by a strategy that predicts $X$ from $Y$, and the conditional entropy. This relation takes the form of an inequality, bounding the error measure in the best case (with the reconstruction strategy that minimizes the error) by an increasing function of the entropy. We note $\hat{x}(Y) \in \X$ the reconstruction model that tries to guess $X$ from $Y$.
    \paragraph{The discrete case}
        In case the input space is discrete, we consider the zero-one reconstruction error for one representation point $Y$: $\varepsilon(Y) = E_X[\1(\hat{x}(Y) \neq X)]$. This is the probability of error when predicting $\hat{x}(Y)$ from $Y$. The function that minimizes the expected error $\mathcal{E} = \E_Y(\varepsilon(Y))$ is $\hat{x}(Y) = \argmax_{x \in \X}p(x\mid Y)$ as shown in Proof 1. We derive the following inequality:
            \begin{align}
                \frac{H(X\mid Y)-1}{\log|\X|}\le \mathcal{E} \le 1 - 2^{-H(X\mid Y)}.
            \end{align}
            
        The left side of the inequality uses Fano's inequality in \cite{element}, the right one is developed in proof 2.
        This first inequalities show how the reconstruction error and the entropy are related. For very invariant representations, it is hard to recover the input from $Y$ and the entropy of $H(X\mid Y)$ is high. 
        
        Besides, there can be an underlying continuity in the input space and it could be unfair to penalize predictions close to the input as much as predictions far from it. We expose another case below that takes this proximity into account.
    \paragraph{The continuous case}
        In the case of convex input space and input variable with continuous density, we consider the 2-norm distance as reconstruction error: $\varepsilon(Y)= \E_{X\mid Y}\big[||X- \hat x(Y)||_2^2\big]$. This error penalizes the average distance of the input and its reconstruction from $Y
        $. The function that minimizes the expected error $\mathcal{E} = \E_Y[\varepsilon(Y)]$ is the conditional expected value: $\hat{x}(y) = \E[X\mid Y=y]$. Then $\mathcal{E} = \var(X\mid Y)$. Helped by the well-known inequality $H(X\mid Y) \le \frac{1}{2}\ln(2 \pi e \var(X\mid Y))$ we obtain:
            \begin{align}
                \frac{e^{2H(X\mid Y)}}{2\pi e} \le \mathcal{E}.
            \end{align}
        Here again, notice that a high entropy $H(X\mid Y)$ implies a high reconstruction error in the best case.
            
    \subsection{Proof 1}
    We have
        \begin{align}
            \mathcal{E} &=   \int_{\Y}p(y)\varepsilon(y)\dy \\
            &=   \int_{\Y}p(y)p(X \neq \hat x(y)\mid y)\dy \\
            &=   \int_{\Y}p(y)(1 - p( \hat x(y)\mid y))\dy.
        \end{align}        
        Since
        \begin{equation}
            p( \hat x(y)\mid y) \le  p(\argmax_{x \in X}p(x\mid y)\mid y),
        \end{equation}
        the reconstruction that minimizes the error is $\hat x(y) = \argmax_{x \in X}p(x\mid y)$.
        However, this is theoretical because in most cases $p(x\mid Y)$ is unknown.
    
    \subsection{Proof 2}
    
    We have:
        \begin{align}
            \log(1- \mathcal{E}) &=  \log\left( \int_{\Y}p(y)(1 -\varepsilon(y)) \dy\right) \\
            & = \log\left( \int_{\Y}p(y)p(\hat x(y)\mid y)\right)\dy \\
            & \ge \int_{\Y}p(y)\log(p(\hat x(y)\mid y))\dy\label{eq:proof2jensen} \\
            & = \int_{\Y}p(y)\int_{\X}p(x\mid y)\log(p( \hat x(y)\mid y)) \dx \dy \\
            & \ge \int_{\Y}p(y)\int_{\X} p(x\mid y)\log(p(x\mid y)) \dx \dy \label{eq:proof2p1} \\
            & = - H(X\mid Y).
        \end{align}
        The Eq.~(\ref{eq:proof2jensen}) is obtained using Jensen inequality and Eq.~(\ref{eq:proof2p1}) is obtained using the result of Proof 1.
        
        Thus,
        \begin{equation}
            \mathcal{E} \le 1 - 2^{-H(X\mid Y)}.
        \end{equation}

\section{SHADE Gradients}
    \label{sec:gradients}
    
    Here is studied the influence of SHADE on a gradient descent step for a single neuron $Y$ of a single layer and for one training sample $X$. The considered case of a linear layer, we have: $Y = \w^\top X + b$.
    
    The gradient of $\Omega_\mathrm{SHADE}$ with respect to $\w$ is:
    \begin{align*}
        \nabla_{\w} \Omega_\mathrm{SHADE} &= (\delta_1 + \delta_2)\x \\
        \text{with } \delta_1 &= \sigma'(y)\big((y-\mu^1)^2 - (y - \mu^0)^2 \big) \\
        \text{and } \delta_2 &= 2\sigma(y)(y - \mu^1) + 2(1-\sigma(y))(y - \mu^0).
    \end{align*}
    With $\sigma(y) = p(Z=1|y)$ which has positive derivative. We can interpret the direction of this gradient by analyzing the two terms $\delta_1$ and $\delta_2$ as follows:
    \begin{itemize}
        \item $\mathbf{\boldsymbol{\delta}_1}$:
        If $(y - \mu^0)^2$ is bigger than $(y -\mu^1)^2$ that means that $y$ is closer to $\mu^1$ than it is to $\mu^0$. Then $\delta_1$ is positive and it contributes to increasing $y$ meaning that it increases the probability of $Z$ being from mode 1. In a way it increases the average margin between positive and negative detections. Note that if there is no ambiguity about the mode of $Z$ meaning that $ \sigma(y)$ or $1 - \sigma(y)$ is very small then this term has negligible effect.
        \item $\mathbf{\boldsymbol{\delta}_2}$:
        This term moves $y$ toward the $\mu^z$ of the mode that presents the bigger probability. This has the effect of concentrating the outputs around their expectancy depending on their mode to get sparser activation.
    \end{itemize}

\end{document}